\documentclass{article}

\usepackage{arxiv}

\usepackage[utf8]{inputenc} 
\usepackage[T1]{fontenc}    
\usepackage{hyperref}       
\usepackage{url}            
\usepackage{booktabs}       
\usepackage{amsfonts}       
\usepackage{nicefrac}       
\usepackage{microtype}      
\usepackage{lipsum}		
\usepackage{graphicx}
\usepackage{natbib}
\usepackage{doi}
\usepackage{amsmath}
\usepackage{amssymb}
\usepackage{amsfonts}
\usepackage{multirow}
\usepackage{amsthm}
\usepackage{ulem}

\usepackage{graphicx}
\usepackage{url} 
\usepackage{latexsym}
\usepackage{float}

\usepackage{booktabs,multirow,mathrsfs} 
\usepackage{titling}
\usepackage{hyperref}
\usepackage{array,color}
\usepackage{caption}
\usepackage{tikz}
\usepackage{subcaption}
\usepackage{epstopdf}
\usepackage{appendix}
\usepackage{listings}

\theoremstyle{definition}
\newtheorem{definition}{Definition}
\newtheorem{theorem}{Theorem}

\title{Hyperparameter Importance for Machine Learning Algorithms}

\author{ {\includegraphics[scale=0.06]{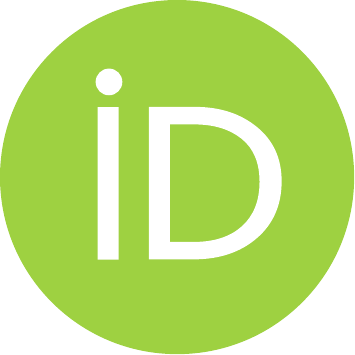}\hspace{1mm}Honghe Jin}\thanks{Corresponding Author.} \\
	Corporate Risk, Wells Fargo, USA\\
	Charlotte, NC 28277 \\
	\texttt{honghe.jin@wellsfargo.com} \\
}

\renewcommand{\headeright}{Technical Report}
\renewcommand{\undertitle}{Technical Report}

\hypersetup{
pdftitle={Hyperparameter Importance for Machine Learning Algorithms},
pdfsubject={},
pdfauthor={Honghe Jin},
pdfkeywords={Hyperparameter, Hyperparameter importance, Large data, Subsampling, Machine Learning},
}

\begin{document}
\date{}
\maketitle

\begin{abstract}
Hyperparameter plays an essential role in the fitting of supervised machine learning algorithms. However, it is computationally expensive to tune all the tunable hyperparameters simultaneously especially for large data sets. In this paper, we give a definition of hyperparameter importance that can be estimated by subsampling procedures. According to the importance, hyperparameters can then be tuned on the entire data set more efficiently. We show theoretically that the proposed importance on subsets of data is consistent with the one on the population data under weak conditions. Numerical experiments show that the proposed importance is consistent and can save a lot of computational resources.
\end{abstract}

\section{Introduction}
\label{sec:intro}
Hyperparameter selection plays an essential role in the fitting of modern supervised machine learning algorithms \citep{james2013introduction, jordan2015machine, mohri2018foundations}. It controls the complexity and topology of the algorithms. Therefore, hyperparameters must be selected carefully before fitting machine learning models to a data set. A data-dependent hyperparameter tuning method is to fit and evaluate models on independent training and test data, such as cross-validation, see, e.g., \cite{stone1978cross}. The hyperparameter searching methods include grid search \citep{lerman1980fitting} and  random search \citep{bergstra2012random}. Alternatively, recent works develop methods that are more specific to certain problems, such as Bayesian optimization \citep{bergstra2011algorithms, hutter2011sequential, klein2017fast}, evolutionary
optimization \citep{loshchilov2016cma}, and meta-learning\citep{brazdil2008metalearning, gomes2012combining, van2015fast}. 

With all the hyperparameters methods above, 
selecting the best hyperparameters for machine learning algorithms usually takes a longer time than fitting a single model. For example, we need to fit $m\times K$ models in a $K$-fold cross-validation tuning procedure to find the best hyperparameters, where $m$ is the number of hyperparameter combination candidates. The more the hyperparameter to be tuned, the larger the number of combinations $m$ is. When the data size is large, fitting $m\times K$ models is almost computationally impossible. We need to either reduce the number of hyperparameters manually or tune them sequentially \citep{wistuba2015sequential}. Then the questions become (1) what exactly are the hyperparameters to be tuned and why; (2) in what order do we tune the hyperparameters sequentially. These two questions require ranking the importance of each hyperparameter for  machine learning algorithms.

In general, people explain the hyperparameter importance based on the understanding of the machine learning algorithms and rank the importance by experience. However, this is not convincing and the hyperparameter importance should not be universal. For example, the maximum depth of a decision tree model should be important when the data has interaction effects, but its importance is negligible if the data comes from an additive model. Therefore, a data-decide hyperparameter importance definition is needed. In \cite{hutter2013identifying}, the researchers identified the most important hyperparameters via forward selection in a greedy search manner. A measure of the hyperparameter importance by a function ANOVA approach was proposed in \cite{hutter2014efficient}. \cite{van2018hyperparameter} studied the significance of hyperparameter across different data sets. Similarly, \cite{probst2019tunability} defined the tunability of hyperparameters by comparing the empirical performance models for multiple data sets. A hyperparameter analysis approach based on Bayes optimization was proposed in \cite{sun2019hyperparameter}.
Other than the general framework for measuring the hyperparameter importance, there are some works targeting some specific machine learning algorithms, such as decision tree \citep{mantovani2016hyper} and neural network \citep{bergstra2012random}.

Although there is some related research studying the hyperparameter importance based on the working data set, to the best of our knowledge, no work has been done to scale the method to a data set when the data size is gigantic. To fill this gap, we proposed a generic definition of  hyperparameter importance that can be estimated through a subsampling procedure. More specifically, we use the variance of risks with varying hyperparameter values to quantify the importance of the hyperparameter, which is proved to be consistent when subsampling a subset from the entire data set. 

The main objectives and proposed approaches in this paper are:
\begin{itemize}
    \item Speed up the hyperparameter tuning: tune the hyperparameters sequentially by the proposed importance; tune the most important hyperparameters by the proposed importance
    \item Rank the hyperparameter importance: empirically estimate the hyperparameter importance by studying the variance of risk with varying hyperparameters
    \item Scale the approach to big data: estimate the importance by subsampling
\end{itemize}


The rest of the article is organized as follows. Section~\ref{sec:method} introduces the definition and theoretical results of the proposed hyperparameter importance. The estimation procedures are provided in Section~\ref{sec:est}. Three case studies are shown in Section~\ref{sec:case}. Section~\ref{sec:discuss} concludes this paper and provides some discussions.
\section{Methodology}
In this section, we provide the definition of the proposed hyperparameter importance and the theoretical results.

\label{sec:method}
\subsection{Notations}
\label{sec:notaion}
Consider a joint distribution $\mathcal{P}$ on $(X, Y)$, where $Y$ is the response variable, and $X$ is an explanatory vector. The data set $\mathcal{T}_n$ comes from the distribution $\mathcal{P}$ with the sample size $n$. Denote the population of $(X, Y)$ as $\mathcal{T}$, i.e., $\mathcal{T} = \lim_{n\to\infty} \mathcal{T}_n$. A machine learning model $f$ with hyperparameters $\boldsymbol{\theta}\in\boldsymbol{\Theta}$ is then fitted to study the relationship between $Y$ and $X$, where $\boldsymbol{\theta} = (\theta_1,...,\theta_q)$, and $\boldsymbol{\Theta} = (\Theta_1,...,\Theta_q)$. We can assume that $\boldsymbol{\theta}$ is uniformly distributed on $\boldsymbol{\Theta}$.
Denote the fitted machine learning model on $\mathcal{T}_n$ as $\hat{f}_n(X;\boldsymbol{\theta})$. With the fitted model, we define a loss function $L(Y, \hat{f}_n(X;\boldsymbol{\theta}))$ and the expected risk $R_n(\boldsymbol{\theta}) = \mathbb{E}_\mathcal{P}[L(Y, \hat{f}_n(X;\boldsymbol{\theta}))]$. The risk $R_n(\boldsymbol{\theta})$ is the expected loss on data $\mathcal{T}_n$ with the fixing hyperparameter $\boldsymbol{\theta}$. Then the best hyperparameter of model $f$ on $\mathcal{T}_n$ is
\begin{equation*}
    \hat{\boldsymbol{\theta}} = \arg\min_{\boldsymbol{\theta}\in\boldsymbol{\Theta}}R_n(\boldsymbol{\theta})
\end{equation*}

In the population version, we define the Bayes risk 
\begin{equation*}
    R^*(\boldsymbol{\theta}) = \inf_{f\in\mathcal{M}} \mathbb{E}_\mathcal{P}[L(Y, f(X;\boldsymbol{\theta}))],
\end{equation*}
and the Bayes optimizer 
\begin{equation*}
    f^* = \arg\inf_{f\in\mathcal{M}} \mathbb{E}_\mathcal{P}[L(Y, f(X;\boldsymbol{\theta}))],
\end{equation*}
where $\mathcal{M}$ is all the possible fitted models in the model space, e.g., all the possible decision trees with possible splits, depths, etc. 

\subsection{Hyperparameter Importance}
\label{sec:def}
We  now   define the hyperparameter importance. 
\begin{definition}
\label{def:im}
Following the notations in Section~\ref{sec:notaion}, the importance of the $j^{\text{th}}$ hyperparameter $\theta_j$ is defined by $\text{Var}_{\theta_j}(R^*(\boldsymbol{\theta}))$, which is the variance of the risk on the hyperparameter $\theta_j$. 
\end{definition}

The intuition behind Definition~\ref{def:im} is that $\text{Var}_{\theta_j}(R^*(\boldsymbol{\theta}))$ measures the variability of the risk when $\theta_j$ is changed. If the risks change much when $\theta_j$ varies, then $\theta_j$ should be tuned carefully since it can influence the results. Otherwise, if the results change only a little with different values of $\theta_j$, that means $\theta_j$ has no significant effect on the model. 

While calculating the variance of risk on $\theta_j$, we need to integral out other hyperparameters $\theta_k$, $k\neq j$. Without loss of generality, we assume there are two hyperparameters $(\theta_1, \theta_2)$ in the model. There are two possible ways to integral out other hyperparameters:
\begin{itemize}
    \item integral out the other hyperparameters before the variance of $\theta_1$
    \begin{equation}
        \int_{\Theta_1}\left(\int_{\Theta_2}R(\theta_1,\theta_2)\,d\theta_2 - \int_{\Theta_1}\int_{\Theta_2}R(\theta_1,\theta_2)\,d\theta_2d\theta_1  \right)^2 \,d\theta_1
        \label{eq:before}
    \end{equation}
    \item integral out the other hyperparameters after the variance of $\theta_1$
    \begin{equation}
        \int_{\Theta_2}\int_{\Theta_1}\left(R(\theta_1,\theta_2) - \int_{\Theta_1}R(\theta_1,\theta_2)\,d\theta_1
        \right)^2\,d\theta_1d\theta_2
        \label{eq:after}
    \end{equation}
\end{itemize} 
Although the values of variances are different when we integral out other hyperparameters, the essential terms in (\ref{eq:before}) and (\ref{eq:after}) are equivalent when we rank the importance of different hyperparameters. That means, we have 
\begin{equation*}
    \text{Var}_{\theta_1}(R(\boldsymbol{\theta})) - \text{Var}_{\theta_2}(R(\boldsymbol{\theta})) = \int_{\Theta_1}\left(\int_{\Theta_2}R(\theta_1,\theta_2)\,d\theta_2 \right)^2 \,d\theta_1 - \int_{\Theta_2}\left(\int_{\Theta_1}R(\theta_1,\theta_2)\,d\theta_1 \right)^2 \,d\theta_2
\end{equation*}
for using either (\ref{eq:before}) or (\ref{eq:after}). Therefore, in the real application cases, we can take an average of the performance for each possible value of a hyperparameter, and then calculate the variance. More details will be shown in Section~\ref{sec:est}.

In addition to the importance of a single hyperparameter, we can also study the effect of hyperparameter combinations. For example, the combination importance of $(\theta_j,\theta_k)$ can be defined as $\text{Var}_{\theta_j,\theta_k}(R^*(\boldsymbol{\theta}))$, in which we calculate the variance of $R^*(\boldsymbol{\theta})$ with varying $(\theta_j,\theta_k)$ and integral out  other hyperparameters. This can help  us study the joint effect of some confounding hyperparameters, such as the number of trees and the step size in the gradient boost models.

\subsection{Consistency of the Hyperparameter Importance Rankings}
\label{sec:consistency}
As was defined in Section~\ref{sec:notaion}, $R^*(\boldsymbol{\theta})$ is the lowest risk of the machine learning model on the population. On the data set $\mathcal{T}_n$, we can decompose the risk by 
\begin{equation*}
    R_n(\boldsymbol{\theta}) = R^*(\boldsymbol{\theta}) + \epsilon_n(\boldsymbol{\theta}),
\end{equation*}
where $\epsilon_n(\boldsymbol{\theta})$ is the error term resulting from the sample. Since the sample case cannot outperform the whole population, and the performance will be better with increasing sample sizes, we can assume $\epsilon_n(\boldsymbol{\theta})\geq 0$ and $\epsilon_n(\boldsymbol{\theta})\to0$ as $n\to\infty$. 

Without loss of generality, we assume there are two hyperparameters $(\theta_1, \theta_2)$ in the model and $\text{Var}_{\theta_1}(R^*(\boldsymbol{\theta}))>\text{Var}_{\theta_2}(R^*(\boldsymbol{\theta}))$. In other words, the hyperparameter $\theta_1$ is more important than $\theta_2$ in the population version. On the data set $\mathcal{T}_n$, we expect the importance rankings are consistent with that in the population version. With some straightforward calculation, we have 
\begin{equation}
    \label{eq:consistency}
    \text{Var}_{\theta_1}(R_n(\boldsymbol{\theta})) - \text{Var}_{\theta_2}(R_n(\boldsymbol{\theta})) = \left[ \text{Var}_{\theta_1}(R^*(\boldsymbol{\theta})) - \text{Var}_{\theta_2}(R^*(\boldsymbol{\theta}))\right] + \left[\text{Var}_{\theta_1}(\epsilon_n(\boldsymbol{\theta})) - \text{Var}_{\theta_2}(\epsilon_n(\boldsymbol{\theta}))\right].
\end{equation}

Since the first term on the right side in (\ref{eq:consistency}) is greater than 0, the condition of hyperparameter importance ranking consistency is 
\begin{equation}
    \label{eq:condition}
    |\text{Var}_{\theta_1}(\epsilon_n(\boldsymbol{\theta})) - \text{Var}_{\theta_2}(\epsilon_n(\boldsymbol{\theta}))| <  |\text{Var}_{\theta_1}(R^*(\boldsymbol{\theta})) - \text{Var}_{\theta_2}(R^*(\boldsymbol{\theta}))|.
\end{equation}
Considering that $\epsilon_n(\boldsymbol{\theta})\to0$ as $n\to\infty$, we can conclude that (\ref{eq:condition}) holds with a large enough sample size $n$.
\begin{theorem}
\label{thm:consistency}
The rankings of sample hyperparameter importance  converge to the rankings of the population hyperparameter importance as $n\to\infty$.
\begin{equation*}
    \lim_{n\to\infty}\text{sign}\left(\text{Var}_{\theta_j}(R_n(\boldsymbol{\theta})) - \text{Var}_{\theta_k}(R_n(\boldsymbol{\theta}))\right) = \text{sign}\left(\text{Var}_{\theta_j}(R^*(\boldsymbol{\theta})) - \text{Var}_{\theta_k}(R^*(\boldsymbol{\theta}))\right), k\neq j
\end{equation*}
\end{theorem}

Suppose we have a large data set $\mathcal{T}_0$ with the sample size $n_0$, which satisfies the condition in (\ref{eq:condition}), we can sample a subset $\mathcal{T}_1$ with a smaller sample size $n_1$ that satisfies (\ref{eq:condition}) as well. Then we can conclude the same hyperparameter importance ranking in the subset $\mathcal{T}_1$ and save the computational time. The hyperparameter tuning on the entire data set $\mathcal{T}_0$ is thus more efficient by either sequentially tuning or most important hyperparameters tuning.
Note that different machine learning algorithms have different convergence speeds, and a lot of them have no explicit formula of the convergence speed. Therefore, there is no general rule to select the best subsample size $n_1$. One way to check the consistency is to select different sizes of subsamples and make sure the importance ranks are identical.
Some examples are shown in Section~\ref{sec:case}.

\subsection{Multiple Subsampling}
In the estimation of the hyperparameter importance, we can sample from the entire data set multiple times to reduce the variance of $\epsilon_n(\boldsymbol{\theta})$. Suppose we sample data sets $\mathcal{T}_{nt}$ with the sample size $n$ for $T$ times, $t = 1,...,T$. Then the condition (\ref{eq:condition}) becomes 
\begin{equation*}
    \label{eq:condition1}
    \left|\text{Var}_{\theta_1}\left(\frac{1}{T}\sum_{t=1}^T\epsilon_{nt}(\boldsymbol{\theta})\right) - \text{Var}_{\theta_2}\left(\frac{1}{T}\sum_{t=1}^T\epsilon_{nt}(\boldsymbol{\theta})\right)\right| <  \left|\text{Var}_{\theta_1}\left(R^*(\boldsymbol{\theta})\right) - \text{Var}_{\theta_2}(R^*\left(\boldsymbol{\theta})\right)\right|.
\end{equation*}
Since 
\begin{equation*}
    \begin{split}
        \text{Var}\left(\frac{1}{T}\sum_{t=1}^T\epsilon_{nt}(\boldsymbol{\theta})\right) &= \frac{1}{T}\text{Var}\left( \epsilon_{n}(\boldsymbol{\theta})\right) + \frac{1}{T^2} \sum_{t\neq s} \text{Cov}\left(\epsilon_{nt}(\boldsymbol{\theta}), \epsilon_{ns}(\boldsymbol{\theta})\right) \\
        &\leq \text{Var}\left( \epsilon_{n}(\boldsymbol{\theta})\right),
    \end{split}
\end{equation*}
the convergence requires a smaller sample size $n$ in Theorem~\ref{thm:consistency}.

Multiple subsampling could increase the estimation time linearly. However, the computational time increases exponentially with increasing sample sizes for some machine learning algorithms. In addition, a smaller sample size requires less computation and storage memory. Therefore, multiple sampling could save computation resources in many applications.
\section{Estimation Procedures}
\label{sec:est}
\begin{figure}
	\centering
\includegraphics[width=0.95\linewidth,trim=.2cm .2cm .2cm 0.1cm,clip]{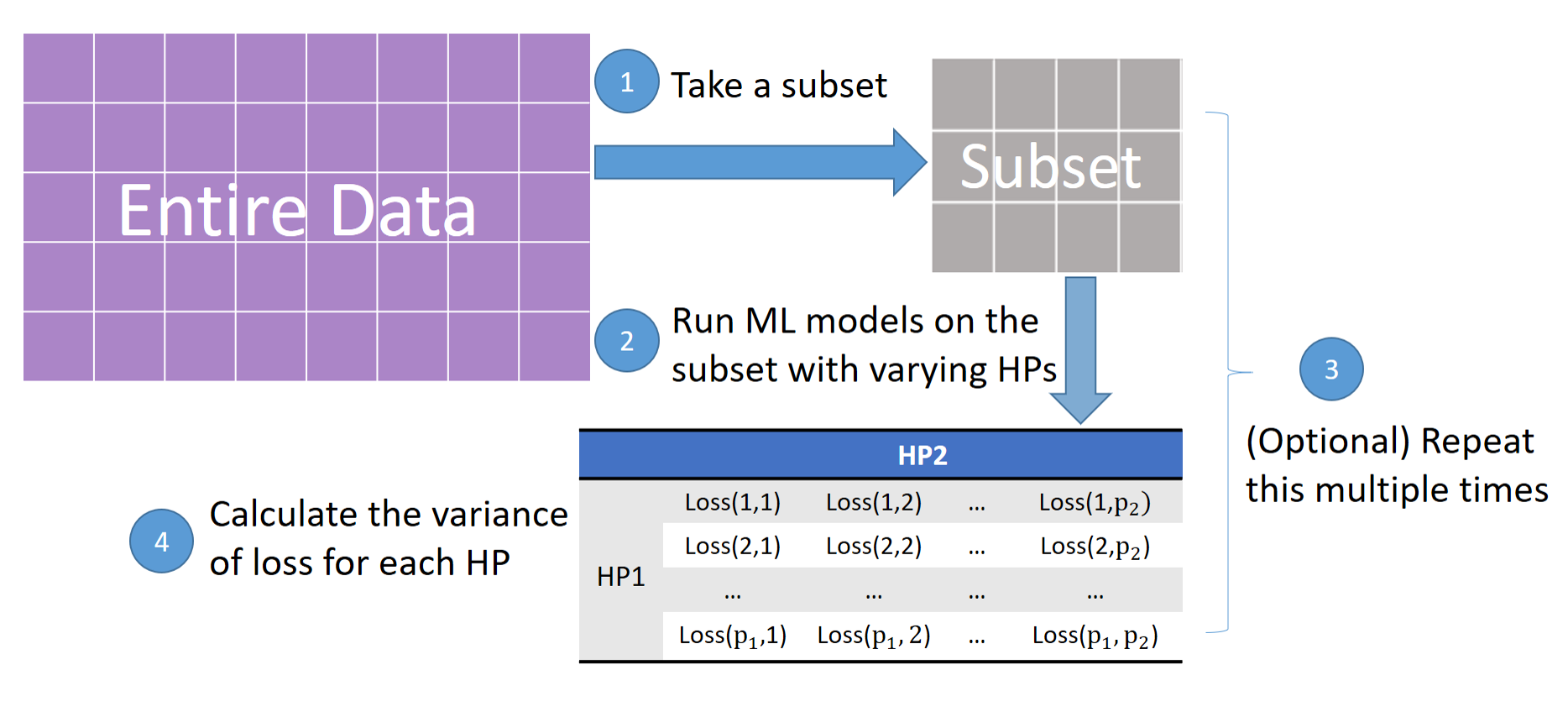}
	\caption{Hyperparameter (HP) importance estimation procedures via subsampling.}
	\label{fig:est}
\end{figure}
In this section, we introduce the details of the proposed hyperparameter importance estimation procedures. Followings are the specific steps of the proposed method.
\begin{enumerate}
    \item Sample a subset from the entire data set
    \item Run grid search on the subset and calculate the loss for each hyperparameter combination
    \item (optional) Repeat the sample procedures for $T$ times and get a $T\times p_1\times...\times p_q$ hyper-matrix for the loss, where $p_1,...,p_q$ are the counts of each candidate hyperparameter
    \item Calculate the average variance of each dimension and estimate the hyperparameter importance
    \item Apply the estimated importance on the entire data set and fit the model more efficiently
\end{enumerate}
A visual illustration of the estimation procedures is shown in Figure~\ref{fig:est}. 

In the implementation, the widely-used range of the parameter space is used to specify the parameter space $\boldsymbol\Theta$. The hyperparameter space for the subsample should be consistent with that to fit the entire data so that the subset could recover the hyperparameter importance for the entire data. A uniform grid of hyperparameter combinations can then be used to get the loss hyper-matrix, as is shown in Figure~\ref{fig:est}. Given the loss hyper-matrix, the importance of each hyperparameter can be estimated by calculating the average of variance for each dimension. For example, there are two hyperparameters in the loss matrix in Figure~\ref{fig:est}. The importance of hyperparameter 1 is estimated by the average of all the column variances. Similarly, the average of row variances estimates the importance of hyperparameter 2.

\section{Case Study}
\label{sec:case}
We test the hyperparameter importance method on three different cases. In particular, we use one public available data and two confidential data sets to demonstrate the advantages of the proposed method. The credit card fraud data fitted by XGBoost shows that the consistency proved in Section~\ref{sec:consistency} holds in a subset with fairly small sample sizes. We use the loan default data to compare the fitting results with/without using the proposed method. The credit card default data set is studied to explain how the proposed method selects the hyperparameters to be tuned and saves computational resources.

\subsection{Credit Card Fraud by XGBoost}
\label{sec:fraud}
In this section, we apply the proposed method on a credit card fraud data set with 284,807 samples. One can download the data from \url{https://www.kaggle.com/mlg-ulb/creditcardfraud}. The data contains a response variable with 28 principal components of the original credit risk data. We perform an XGBoost model on the data to do a binary classification and study the hyperparameters importance. The hyperparameters of interest are Max Depth, Step Size, Max Iteration, Subsample Rate, Column Sample Rate,  Regularizations $\alpha,\lambda$, and Minimum Loss Reduction $\gamma$. In addition to the single hyperparameter importance, we also check the joint importance of Step Size and Max Iteration since they have a confounding effect.

We split the data into a training set (70\%) and a test set (30\%). In each iteration of the estimation procedures, we select a subset from the training set and evaluate it on the test set. The performance is quantified by the area under the curve (AUC). We sample the data with the size = 2000, 5000, 7000 to check the consistency. The subsampling is repeated for 10 times. 

\begin{figure}
	\centering
\includegraphics[width=1\linewidth,trim=.2cm .2cm .2cm 0.1cm,clip]{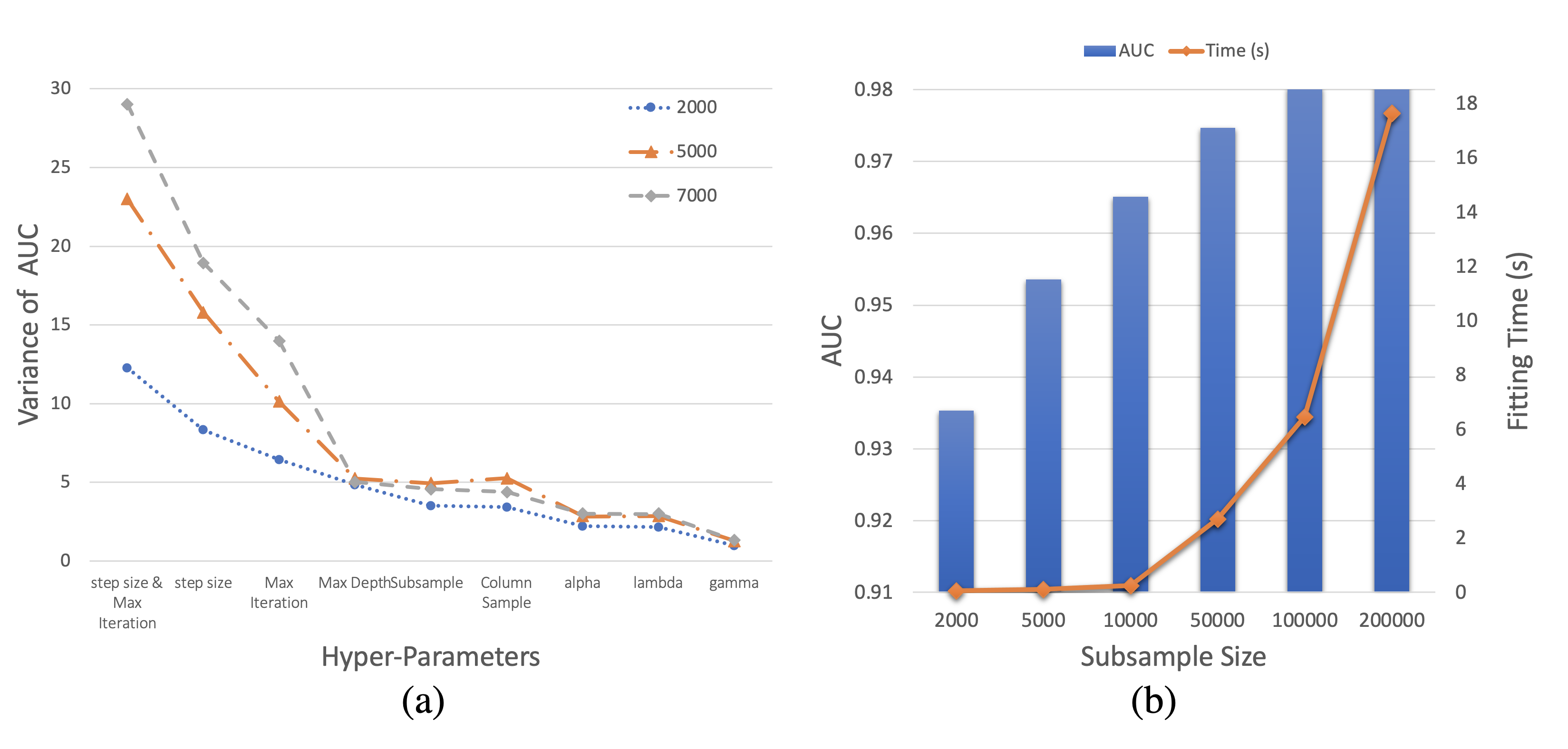}
	\caption{The results for the credit card fraud data: (a) variances of AUC ($\times 10^{-6}$) for each hyperparameter when the subsample size = 2000, 5000, 7000; (b) AUC and fitting time with different subsample sizes.}
	\label{fig:fraud}
\end{figure}

The variances of AUC ($\times10^{-6}$) for the hyperparameters are shown in Figure~\ref{fig:fraud}, panel (a). From the result, we conclude that the most important hyperparameter is the Step Size, followed by Max Iteration, Max Depth, Subsample Rate, Column Sample Rate, $\alpha$, $\lambda$, and $\gamma$. Since the importance rankings are consistent with different subsample sizes, the hyperparameter importance rankings in Figure~\ref{fig:fraud} should recover the correct rankings on the entire data set. We can then apply the result on the entire data set to tune the hyperparameters more efficiently. We show how the hyperparameter importance improves the efficiency of hyperparameter tuning in Section~\ref{sec:loan}, in which the data set is smaller and can be tuned on a more dense hyperparameter grid.

In addition, we show the effectiveness of the of the subsampling  procedures in this case study. The average AUC and fitting time in seconds for different subsample sizes are presented in Figure~\ref{fig:fraud}, panel (b). It is observed that the AUC starts converging when the subsample size reaches 50,000. On the other hand, the fitting time grows exponentially as the sample size increases. Therefore, the subsampling procedures save a considerable amount of computational resources for large data sets.

\subsection{Loan Default by XGBoost}
\label{sec:loan}
The data set contains about 21,305 customers' information with a target variable $Y$ indicating their loan defaults $(Y = 1)$ or not $(Y=0)$ from the Wells Fargo Bank. There are 19 explanatory features in the data. We apply the XGBoost model to fit a binary classification model to predict the default. The tuned hyperparameter candidates are the same as those in Section~\ref{sec:fraud}. Similarly, we split the data into a training set (70\%) and a test set (30\%). We repeat the subsampling  10 times with the sample size 3000 and 5000.

The hyperparameters from the most important to the least important are
Max Depth, Step Size, Max Iteration, $\alpha$, $\lambda$, Column Sample Rate, Subsample Rate, and $\gamma$. The orders, in this case, is different from those in Section~\ref{sec:fraud}, which confirms that the hyperparameter importance is different across different cases. Considering the sample size of this data set, we are able to tune the hyperparameters by grid search on the entire data set.  We can thus explore how the hyperparameter importance can improve the hyperparameter tuning efficiency.

With the estimated hyperparameter importance, we tune the hyperparameters sequentially by group with the importance rankings. In particular, we first tune the hyperparameters (Max Depth, Step Size, Max Iteration), and then tune ($\alpha$, $\lambda$), and then tune (Column Sample Rate, Subsample Rate), and tune the final hyperparameter $\gamma$. The untuned hyperparameters are fixed as the default values. As a benchmark, we grid search all the hyperparameters simultaneously on the entire data set. To have a fair comparison, the grids of each hyperparameter are the same for the simultaneously tuning and the sequentially tuning, and both methods are run on the same device.

It takes 697.12 minutes to finish the grid search of all the hyperparameters on the entire data set. The AUC on the test data is 0.8713. In the sequential tuning, it takes 5.36 minutes to select (Max Depth, Step Size, Max Iteration), 4.15 minutes to tune ($\alpha$, $\lambda$), 3.32 minutes to tune (Column Sample Rate, Subsample Rate), and 0.74 minutes to tune $\gamma$. 
The test AUC by sequential tuning is 0.8711.  The best hyperparameters selected by the two tuning methods are shown in Table~\ref{tab:hps}. The sequential hyperparameter tuning with the estimated importance yields similar selected hyperparameters and AUC. Therefore, applying the proposed importance can provide comparably good results and save over 90\% of the computation time in this case study even considering the importance estimation time, which takes 24.21 minutes to run the grid search on subsets.

\begin{table}[h]
	\caption{Selected hyperparameters for the XGBoost model on the credit card fraud data}
	\centering
	\begin{tabular}{l|llllllll}
		\hline\hline
		Method & Max Depth& Step Size& Max Iteration& $\alpha$& $\lambda$& ColumnSample& Subsample& $\gamma$\\\hline
		Simultaneous& 4& 0.15& 150& 0.5& 1.0& 0.3& 1.0& 0.2 \\
		Sequential& 4 & 0.15& 150 & 1.0 & 1.0 & 0.5 & 1.0 & 0.2\\
		\hline
	\end{tabular}
	\label{tab:hps}
\end{table}

\subsection{Credit Card Default by Gradient Boosting}
In this section, we consider a credit card default data set from Wells Fargo Bank of the seasoned low FICO segment. The data contains 50,356,580 samples, 15 explanatory features, and one response variable. Similar to the previous cases, we split it to a training set (60\%) and a test set (40\%) to validate the performance.
We use the gradient boosting machine (GBM) to predict the contractual default of each customer. The hyperparameters being studied are Max Depth, Step Size, Max Iteration, Subsample Rate, Max Bins, and Min Instance Per Node. We sample subsets from the entire data set with the fraction = 0.5\%, 1\%, 2\% for 5 times. AUC is used to quantify the performance.

\begin{figure}
	\centering
\includegraphics[width=1\linewidth,trim=.2cm .2cm .2cm 0.1cm,clip]{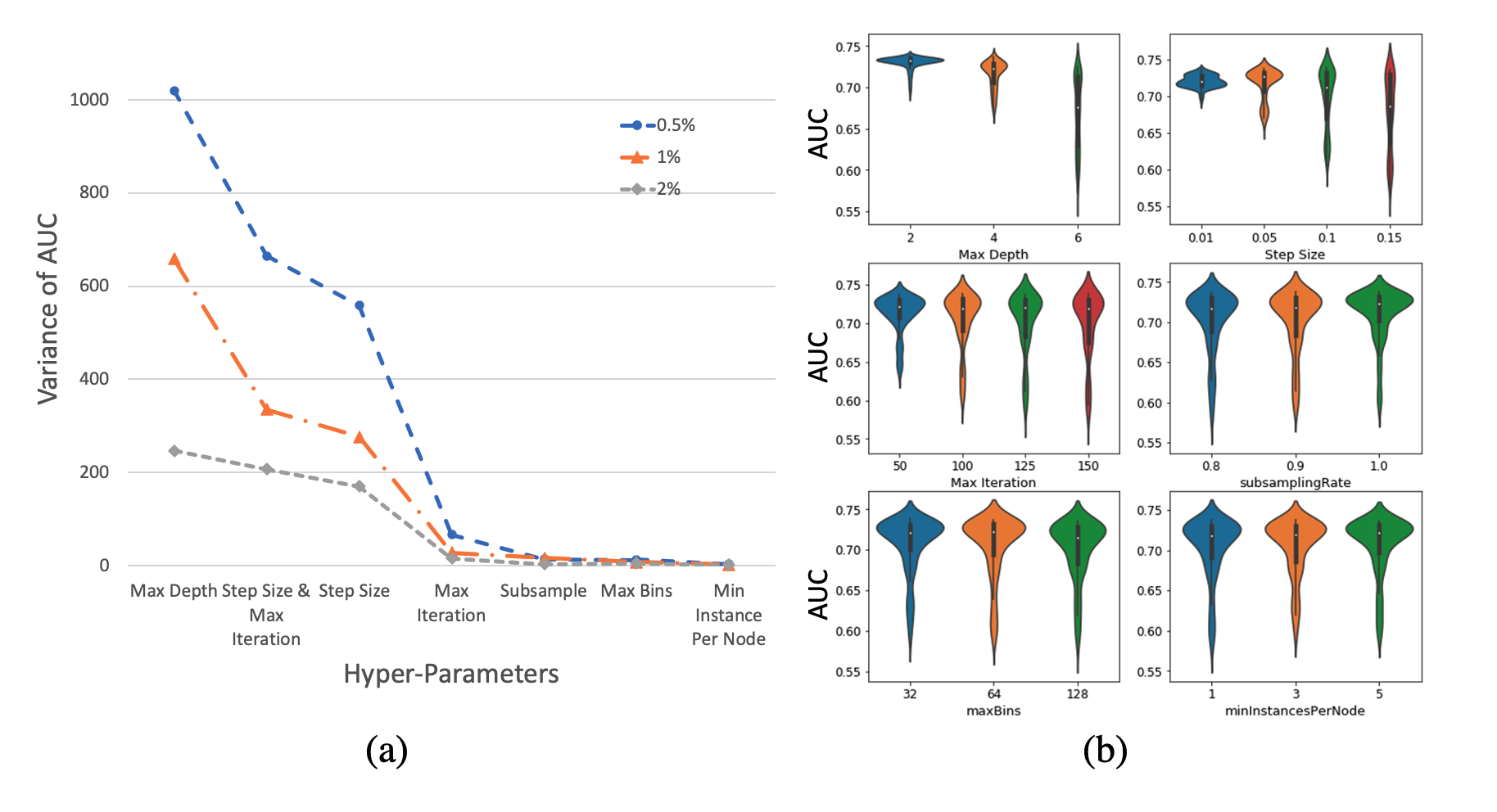}
	\caption{The results for the credit card default data: (a) variances of AUC ($\times 10^{-6}$) for each hyperparameter when the subsample fraction = 0.5\%, 1\%, 2\%; (b) Distributions of AUC for hyperparameters when the subsample fraction = 1\%.}
	\label{fig:default}
\end{figure}

The results are shown in Figure~\ref{fig:default}. Panel (a) shows the importance of AUC for hyperparameters with different subsample fractions. It is clear that the three fractions yield the same importance rankings. The consistency is then demonstrated. We can visualize the distributions of AUC for the grid values of each hyperparameter in panel (b), Figure~\ref{fig:default}. It can explain how the varying hyperparameters change the AUC. For example, in the top left figure, the distributions of AUC are totally different when the Max Depth varies from 2 to 6. This implies that Max Depth must be tuned carefully to get a good result. In the bottom left figure, the shapes of AUC distributions are almost identical when Max Bins is changed. Therefore, the hyperparameter Max Bins is not important in this case study.

Since there is a big jump in the importance trend line in Figure~\ref{fig:default}, panel (a), we can only tune the most important hyperparameters on the entire data so as to save computational resources. In this case, we only need to tune Max Depth, Step Size, and Max Iteration. We tune Max Iteration here because the combined effect of Step Size \& Max Iteration is the second most important. If the computation resources are very limited, tuning only Max Depth and Step Size should give a fairly good result as well. 

In some scenarios, modelers are required to argue why certain hyperparameters are tuned instead of others. Instead of using experience, Figure~\ref{fig:default}, panel (a) should provide a convincing evidence for the tunable hyperparameters selection.

\section{Discussion}
\label{sec:discuss}
In this paper, we proposed a definition of hyperparameter importance by the variance of the risks that can be estimated via subsampling procedures.  This is a generic definition for all the machine learning algorithms with multiple hyperparameters. The hyperparameter tuning can be more efficient by studying the defined hyperparameter importance. 

The subsampling procedures require much less storage memory and computational resources. Therefore, the defined hyperparameter importance can be applied on gigantic data sets. Theoretical analysis and empirical studies demonstrate the consistency of the importance estimated by subsampling. A future work to follow up this paper is to study the convergence speed of the hyperparameter importance for some specific machine learning algorithms instead of the general conclusion for all machine learning algorithms. 

\section*{Acknowledgement}
We are grateful to Sheri Kong for useful suggestions and Shuguang Zhang for his help of data processing in the case studies.

\newpage

\bibliographystyle{unsrtnat}
\bibliography{main}  

\end{document}